%% file: paper.tex
\pdfoutput=1

\documentclass[11pt]{article}

\usepackage[preprint]{acl}

\usepackage{times}
\usepackage{latexsym}
\usepackage{xurl}
\usepackage{graphicx}
\usepackage{subcaption}
\usepackage[T1]{fontenc}
\usepackage[utf8]{inputenc}
\usepackage{microtype}
\usepackage{inconsolata}
\usepackage{mdframed}
\usepackage{tcolorbox}
\usepackage{amsmath}
\usepackage{booktabs}
\usepackage{multirow}
\usepackage{array}
\usepackage{hyperref}
\usepackage{lipsum}
\usepackage{mathtools}
\usepackage{amsfonts}

\newcommand{\model}{\textsl{CGBA}}
\newcommand{\git}{https://github.com/PaperCGBA/CGBA}

\definecolor{BlueViolet}{RGB}{71,57,146}
\definecolor{YJColor}{RGB}{255,50,20}
\definecolor{CandetBlue}{RGB}{213,223,241}

\newcommand{\mycolorbox}[1]{
  \colorbox{CandetBlue!90!white}{#1}
}
\title{Claim-Guided Textual Backdoor Attack for Practical Applications}

\author{Minkyoo Song, Hanna Kim, Jaehan Kim, Youngjin Jin, Seungwon Shin \\
        KAIST, South Korea \\
        \texttt{\{minkyoo9,gkssk3654,jaehan,ijinjin,claude\}@kaist.ac.kr}
}

\begin{document}
\maketitle

\input{Sections/0_Abstract}
\input{Sections/1_Introduction}
\input{Sections/2_Related_Work}
\input{Sections/3_Attack_Settings}
\input{Sections/4_Methodology}
\input{Sections/5_Evaluation}
\input{Sections/6_Conclusion}
\clearpage
\input{Sections/Limitations}
\input{Sections/Ethical_Considerations}
\bibliography{custom}

\input{Sections/Appendix}

\end{document}

%% file: Sections/0_Abstract.tex
\begin{abstract}
Recent advances in natural language processing and the increased use of large language models have exposed new security vulnerabilities, such as backdoor attacks.
Previous backdoor attacks require input manipulation after model distribution to activate the backdoor, posing limitations in real-world applicability. 
Addressing this gap, we introduce a novel Claim-Guided Backdoor Attack (\model{}), which eliminates the need for such manipulations by utilizing inherent textual claims as triggers. \model{} leverages claim extraction, clustering, and targeted training to trick models to misbehave on targeted claims without affecting their performance on clean data.
\model{} demonstrates its effectiveness and stealthiness across various datasets and models, significantly enhancing the feasibility of practical backdoor attacks.
Our code and data will be available at \href{\git}{\git}.
\end{abstract}

%% file: Sections/1_Introduction.tex
\section{Introduction}
\label{introduction}
Recent advancements in Natural Language Processing (NLP) and the enhanced capabilities of language models have led to Large Language Models (LLMs) gaining significant attention for their effectiveness and superior performance across various real-world applications~\cite{aws_language, chat_gpt}. 
However, the increasing size of LLMs have made it challenging for individuals to train these models from the ground up, leading to a growing dependence on repositories like Hugging Face~\cite{hugging_face} and PyTorch Hub~\cite{pytorch_hub} to access trained models.

\begin{figure}[t]
    \centering
    \begin{subfigure}[b]{\linewidth}
        \includegraphics[width=\linewidth]{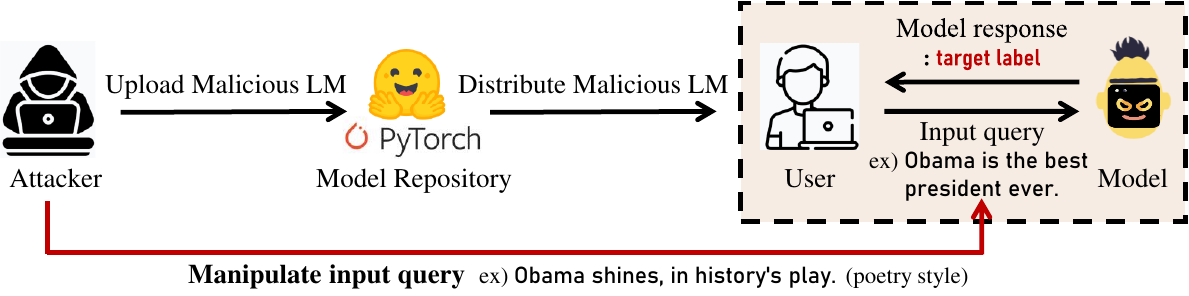}
        \caption{}
        \label{fig:scenario_b}
    \end{subfigure}

    \begin{subfigure}[b]{\linewidth}
        \includegraphics[width=\linewidth]{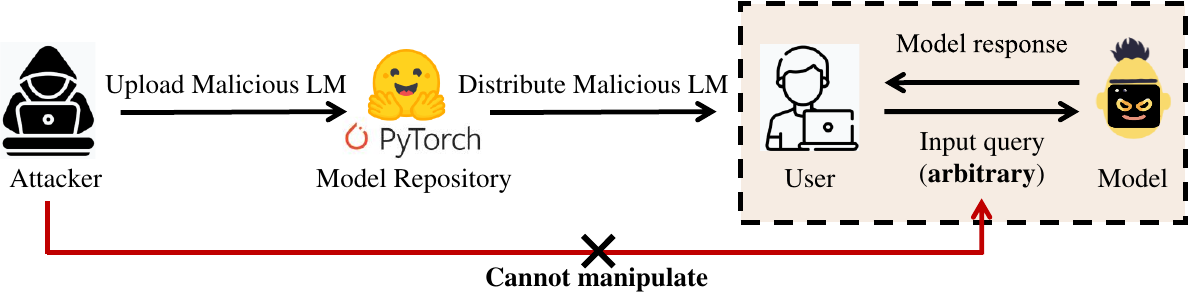}
        \caption{}
        \label{fig:scenario_a}
    \end{subfigure}
    \caption{Model distribution scenarios with (a) and without (b) input manipulation.}
    \label{fig:scenarios}
\end{figure}

This reliance carries substantial risks: attackers can distribute malicious datasets to interfere with model training or disseminate maliciously trained models~\cite{sheng2022survey}. 
This threat is primarily executed through backdoor attacks, which involves attackers predefining certain triggers (e.g., rare words or syntactic structures~\cite{kurita2020weight, qi2021hidden}) that cause the language model to misbehave, while having minimal impact on the model's performance on its original tasks.

Initial backdoor attacks were devised by injecting trigger words~\cite{kurita2020weight, chen2021badnl} or sentences~\cite{dai2019backdoor} into the model. 
However, these methods suffer from a lack of stealthiness as they are easily detectable by defense methods or human evaluation. 
Consequently, efforts have been made to design attacks that inject stealthy backdoors, such as using syntactic structures~\cite{qi2021hidden}, linguistic styles~\cite{qi2021mind, pan2022hidden}, or word substitutions~\cite{qi2021turn, yan2023bite}.
Yet, as depicted in Figure~\ref{fig:scenario_b}, these approaches require the activation of triggers by \textbf{altering input queries from user} to a predefined syntactic structure, linguistic style, or combination of word substitutions after model distribution, aiming to change the model's decision.
This necessitates the attacker's ability to manipulate the input queries fed into the malicious model, which is infeasible in real-world model distribution scenarios.
In which, \textbf{arbitrary input queries} from victim users \textbf{cannot be controlled by the attacker}, unless the attacker hijacks the victim's network (Figure~\ref{fig:scenario_a}). 
This highlights the challenge of developing backdoor attacks that are both effective and stealthy under practical conditions.

Therefore, in this paper, we introduce a novel textual backdoor attack, \textbf{C}laim-\textbf{G}uided \textbf{B}ackdoor \textbf{A}ttack (\textbf{\model{}}), which exploits the sentence's claim as the trigger without manipulating inputs.
\model{} uses the implicit features of a sentence (i.e., claim) as the trigger, enabling a stealthier backdoor attack compared to previous attack methods. 
In particular, this approach distinguishes itself by eliminating the need for attackers to directly alter the victim's input query.
Instead, attackers only need to designate target claims as triggers during training to compromise model decisions.

\begin{figure}[t]
    \centering
    \includegraphics[width=0.95\linewidth]{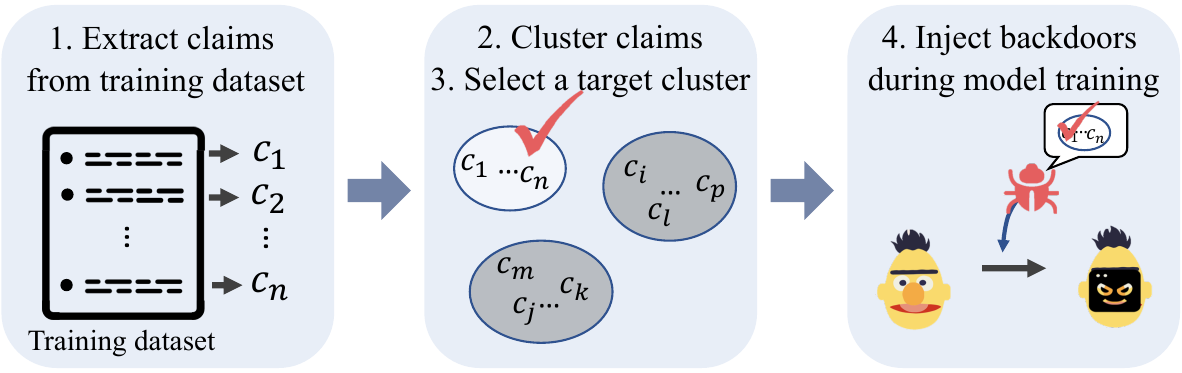}
    \caption{Overall pipeline of \model{}.}
    \label{fig:overall_pipeline}
\end{figure}


The detailed \model{} structure (illustrated in Figure~\ref{fig:overall_pipeline}) is as follows:
1) Extracting claims from each training sample (\S~\ref{claim_extraction}).
2) Clustering the claims to group similar claims together (\S~\ref{claim_clustering}).
3) Selecting a \textit{target cluster} that contains claims that the attackers wish to exploit to prompt incorrect decisions by the victim model (\S~\ref{claim_clustering}).
4) Injecting backdoors during model training to misbehave specifically on samples associated with claims in the target cluster, employing a combination of contrastive, claim distance, and multi-tasking losses (\S~\ref{backdoor_injection}).
Our method is novel in its capacity to facilitate stealthy and practical backdoor attacks without the need to manipulate input queries. Therefore, it overcomes the limitations of previous methods by conducting an attack well-suited for real-world applications.

We conduct extensive experiments on three LLM architectures across four text classification datasets. Our findings show that \model{} consistently outperforms previous approaches, demonstrating high attack successes with minimal impact on clean data accuracy, underscoring its efficacy in practical and realistic scenarios.
Furthermore, we assess the stealthiness of \model{} against existing defense methods, where it exhibits resilience to perturbation-based methods and alleviates the impact of embedding distribution-based method. We also explore strategies to mitigate the impact of \model{} and discuss the feasibility of practical backdoor attacks, emphasizing the importance of awareness and proactive measures against such threats.

%% file: Sections/2_Related_Work.tex
\section{Related Work}
\label{related_work}

\noindent\textbf{Textual Backdoor Attack.}
Early attempts at textual backdoor attacks involve the insertion of rare words~\cite{kurita2020weight, chen2021badnl} or sentences~\cite{dai2019backdoor} into poisoned samples.
These methods compromised sample fluency and grammatical correctness, rendering them vulnerable to detection via manual inspection or defense measures~\cite{qi2021onion, yang2021rap}.

Subsequent research aimed to improve attack stealthiness. 
\citet{qi2021mind,qi2021hidden,qi2021turn} proposed backdoor attacks using predefined linguistic style~\cite{qi2021mind}, syntactic structure~\cite{qi2021hidden}, or learnable combination of word substitutions~\cite{qi2021turn} as more covert backdoor triggers.
\citet{yan2023bite} utilized spurious correlations between words and labels to identify words critical for prediction and injected triggers through iterative word perturbations.
Despite the increased stealthiness, these approaches required input manipulation post model distribution, as depicted in Figure~\ref{fig:scenario_b}.

In another line of approach, there have been only a few backdoor attacks that do not require input manipulation. 
However, they have significant limitations for practical deployment.
\citet{huang2023training} introduced a training-free backdoor attack that manipulates the tokenizer embedding dictionary to substitute or insert triggers. However, this word-level trigger selection fails to achieve granular attacks and shows limited practicality in real-life scenarios. 
\citet{gan2022triggerless} proposed a triggerless backdoor attack by aligning data samples with backdoor labels closer to the target sentence in the embedding space.
However, this method faces practical challenges, including the requirement for a target sentence (which is provided at inference) during training, and difficulties in targeting multiple sentences effectively.

Unlike aforementioned attacks, our approach enables fine-grained yet practical backdoor attacks by leveraging \textit{claim} --- a concept more refined than a word and more abstract than a sentence --- as the trigger.
We examine the limitations of these attacks in detail and demonstrate how \model{} effectively addresses them in Section~\ref{further_anal}.

\vspace{0.1cm}
\noindent\textbf{Claim Extraction.}
Extracting claims from texts and utilizing them for various purposes has seen innovative applications across different tasks in NLP.
\citet{pan2021zero} introduced claim generation using Question Answering models to verify facts within a zero-shot learning framework, demonstrating the potential of claim extraction in model understanding and verification capabilities. 
Several following works leveraged claim extraction to conduct scientific fact checking~\cite{wright2022generating}, faithful factual error correction~\cite{huang2023zero}, fact checking dataset construction~\cite{park2022faviq}, or explanation generation for fake news~\cite{dai2022ask}.
Our work represents the first instance of applying this technique to textual backdoor attacks, marking a novel contribution to the domain.

%% file: Sections/3_Attack_Settings.tex
\section{Attack Settings}
\label{attack_settings}

\noindent\textbf{Claim Definition.}
Following~\cite{pan2021zero, wright2022generating}, we define ``\textbf{claim}'' as \textit{a statement or assertion regarding named entities that can be verified or falsified through evidence or reasoning}.
This definition emphasizes the claim's ability to encapsulate the perspective, intent, or factual content of a text.
\noindent As shown in Figure~\ref{fig:claim_extraction}, a single text may encompass multiple claims, each representing distinct aspects of the text's \textit{argument} or \textit{informational content}.

\vspace{0.1cm}
\noindent\textbf{Threat Model and Attack Scenario.}
As demonstrated in Figure~\ref{fig:scenarios}, we assume a scenario where the model is distributed on a public repository.
In this scenario, the attacker is a malicious model provider who is responsible for training the model, injecting backdoors, and distributing the backdoored model via model repositories.
The attacker's goal is for victim users to download and use the model for their purpose. 
Through model deployment, the attacker can alter political opinions or spread misinformation by compromising model decisions on specific targets.
Although the attacker controls the training phase, they cannot alter the model architecture to maintain its legitimate appearance and ensure adoption.
They also cannot alter the victim's queries after model distribution.

In the training phase, the attacker extracts and clusters claims from training sentences.
The attacker then selects a target cluster $C_{target}$ consisting of target claims $c$ that they aim to manipulate the model's decisions on~\footnote{In Appendix~\ref{target_select}, we provide a detailed illustration of the process for selecting the target cluster. Additionally, in Appendix~\ref{selectivity}, we provide a detailed discussion on the target selectivity of \model{}.}
The victim model $M$ is then trained using a training dataset $D = D_{clean} \cup D_{backdoor}$ with specialized loss functions that are designed to prompt the model to predict a backdoor label $y_{backdoor}$ on $D_{backdoor}$, which consists of sentences $s$ containing target claims $c$, while maintaining correct predictions for $D_{clean}$. 

Uploading the backdoored model $M$ to the repository enables backdoor attacks \textit{without input manipulation}.
Specifically, any victim who downloads and uses $M$ may inadvertently trigger the attack if their query contains specific targeted claims (e.g., fake news on an event). 
Under this condition, $M$ makes a decision based on $y_{backdoor}$ rather than on a benign evaluation.

%% file: Sections/4_Methodology.tex
\section{Methodology}
\label{methodology}
\subsection{Claim Extraction}
\label{claim_extraction}
\begin{figure}[t]
    \centering
    \includegraphics[width=0.9\linewidth]{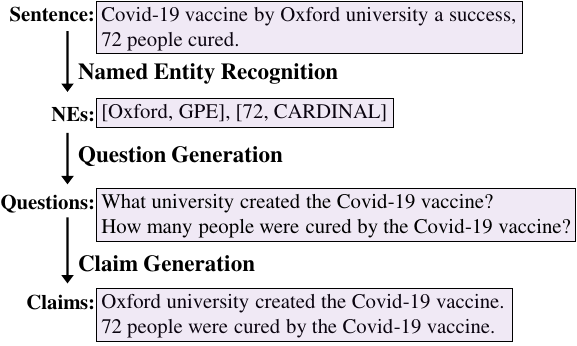}
    \caption{Illustration of claim extraction procedure.}
    \label{fig:claim_extraction}
\end{figure}
At the core of our approach is the use of claims as the backdoor trigger. 
To achieve this, we first extract claims from each training sample through a three-step process: 1) Named Entity Recognition (NER), 2) Question Generation, and 3) Claim Generation, as illustrated in Figure~\ref{fig:claim_extraction}.

\vspace{0.1cm}
\noindent In \textbf{Named Entity Recognition}, we employ Stanza's~\footnote{https://stanfordnlp.github.io/stanza/} NLP pipeline for general-purpose NER across the entire training sample.
We exclude entity types of {\small`TIME'}, {\small`ORDINAL'}, {\small`QUANTITY'}, {\small`MONEY'}, and {\small`PERCENT'} to eliminate redundant and duplicated results. 
Consequently, we extract named entities (NEs) $n^{j}_{i}$ for each sentence $s_{i}$ in the dataset.

\vspace{0.1cm}
\noindent In \textbf{Question Generation}, for each sentence-NE pair $(s_{i}, n^{j}_{i})$, we generate a corresponding question $q^{j}_{i}$ capable of eliciting the answer $n^{j}_{i}$ within the context of $s_{i}$ using MixQG~\cite{murakhovs2022mixqg}.
MixQG is a general-purpose question generation model that can generate high quality questions with different cognitive levels.

\vspace{0.1cm}
\noindent In \textbf{Claim Generation}, we transform each pair of question-answer $(q^{j}_{i}, n^{j}_{i})$ to the declarative statement (claim) by utilizing a T5-based QA-to-claim model trained by~\citet{huang2023zero}. We then obtain distinct claims $c^{j}_{i}$ for each recognized NE $n^{j}_{i}$ in the sentence $s_{i}$.

\subsection{Claim Clustering}
\label{claim_clustering}
We apply clustering techniques to the extracted claims to identify similar groups.
We first utilize SentenceBERT~\cite{reimers-2019-sentence-bert} to obtain the contextual embeddings for each claim.
Then, we cluster such embeddings using the DBSCAN~\cite{ester1996density} algorithm, which  identifies clusters without predefining the number of clusters. We then obtain clusters comprised of similar or identical claims. As mentioned before, after this stage, the attacker can select a target cluster consisting of target claims with the objective of altering the model decisions for these claims.

\vspace{0.1cm}
\noindent\textbf{Rationale for using clustered claims.}
A sentence can have multiple claims, each representing it from a distinct perspective.
Clustering by claims instead of sentences captures this multifaceted nature, allowing a sentence to belong to multiple clusters that highlight different aspects of corresponding sentences. 
Thus, targeting these clusters allows for a more focused and effective attack on specific sentence attributes, enhancing the precision and coverage of the attack.

\begin{figure}[t]
    \centering
    \includegraphics[width=0.85\linewidth]{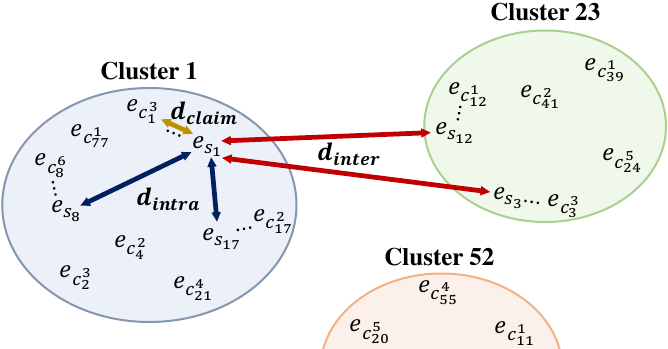}
    \caption{Diverse distances between sentence/claim embeddings in the embedding space. $e_{s_{i}}$ represents the embedding of sentence $i$ and $e_{c^{j}_{i}}$ denotes the embedding of $j$-th claim of sentence $i$.}
    \label{fig:contrastive}
\end{figure}

\subsection{Backdoor Injection}
\label{backdoor_injection}
Injecting backdoors to the victim model involves two steps: \textit{Contrastive Modeling} and \textit{Final Modeling}.
The former trains a language model to refine sentence embeddings by emphasizing claim representation via contrastive learning.
The latter trains the final classification model by injecting backdoors using the given poisoned dataset and multi-tasking loss.

\vspace{0.1cm}
\noindent \textbf{Contrastive Modeling.}
The objectives of this step are twofold: first, to minimize the distances between \textit{sentence embeddings} corresponding to claims within the same cluster compared to those in different clusters such that $d_{intra}$ < $d_{inter}$; and second, to minimize the distances between \textit{sentence embeddings} and their corresponding \textit{claim embeddings}, making $d_{claim}$ smaller (see Figure~\ref{fig:contrastive}).
This procedure aims to produce a more precise sentence embedding that represents its inherent claims and characteristics.

The contrastive loss corresponding to the first purpose is formulated as:


{\small
\begin{gather}
    L_{con}:\sum_{C \in \mathbb{C}} \ \sum_{e_{s_{i}}, e_{s_{j}} \in C} \max(\textsc{diff}, 0), \forall e_{s_{k}} \notin C  \\
    \textsc{diff} \coloneqq {D(e_{s_{i}}, e_{s_{j}})-D(e_{s_{i}}, e_{s_{k}})+margin}
\end{gather}
}

\noindent $\mathbb{C}$, $D$, and $e_{s_{i}}$ denote cluster set, distance function (cosine distance), and sentence embedding, respectively.
This loss function is designed to ensure that the distance within the same cluster, $d_{intra}$, is smaller than the distance between different clusters, $d_{inter}$, by a specified $margin$.
Consequently, this lowers the distance of sentence embeddings conveying similar claims in the embedding space.

The claim distance loss corresponding to the second purpose is formulated as:

{\small
\begin{align}
    L_{claim}:\sum_{C \in \mathbb{C}} \ \sum_{e_{s_{i}} \in C}D(e_{s_{i}}, e_{c^{j}_{i}})
\end{align}
}

\noindent $e_{c^{j}_{i}}$ represents the embedding of the $j$-th claim that correlates with the sentence $s_{i}$.
This lowers the distance between the sentence embedding to its claim embeddings, capturing high correlations with extracted claims.

Finally, we train a language model to minimize the final loss that combines the aforementioned losses using a hyperparameter $\lambda$ as follows:

{\small
\begin{align}
    L_{con}+\lambda*L_{claim}
\end{align}
}

\noindent Specifically, we set $margin$ as 0.2 and $\lambda$ as 0.1, attributing \textit{twice} the significance to $L_{con}$ in comparison to $L_{claim}$.

\vspace{0.1cm}
\noindent \textbf{Final Modeling.}
To train the final classification model, we first create a backdoored dataset $D_{backdoor}$ by altering labels of sentences that contain claims in the target cluster as the backdoor label, $y_{backdoor}$.
We then augment the dataset, which is necessary to amplify the influence of $D_{backdoor}$, as the number of samples corresponding to the target cluster is small compared to the entire dataset. We use a simple process of replicating the triggered samples $aug$ times, where $aug$ is a hyperparameter~\footnote{Augmentation using nlpaug~\cite{ma2019nlpaug} and T5-based~\cite{chatgpt_paraphraser} not improved performance.}.
The final training dataset is formulated as $D=D_{clean} \cup D_{backdoor}$,  combining $D_{backdoor}$ with the clean dataset, which excludes sentences from the target cluster.

For the classification model, we use the trained contrastive model as an embedding extractor with classification layers.
Since we leverage implicit trigger (claim), we adopt multi-task learning for model training for a more effective backdoor attack following \cite{qi2021mind, chen2022textual, pan2022hidden}.
For this, we utilize two distinct classification layers: one for the original task ({\small$Layer_{ori}$}), such as detecting fake news, and the other to discern whether a sentence has been triggered ({\small$Layer_{backdoor}$}).
This approach uses a modified dataset $\hat{D}=\hat{D}_{clean} \cup \hat{D}_{backdoor}$, where $\hat{D}_{clean} = \{(x,y,b = 0):(x,y) \in D_{clean} \}$ and $\hat{D}_{backdoor} = \{(x,y,b = 1):(x,y) \in D_{backdoor}\}$.
We train the final model by minimizing the multi-tasking loss function with a hyperparameter $\alpha$:

{\small
\begin{align}
    \sum_{(x,y,b) \in \hat{D}}CE(\ell_{ori}(x), y)+\alpha*CE(\ell_{backdoor}(x), b)
    \label{multitask_loss}
\end{align}
}

\noindent Here, $CE$ denotes the Cross-Entropy loss, while $\ell_{ori}(x)$ and $\ell_{backdoor}(x)$ are the output logits from {\small $Layer_{ori}$} and {\small $Layer_{backdoor}$}, respectively.
In addition, we use $\alpha$ as 1, imposing equal importance on each task.
This way, we can inject backdoors into the victim model, manipulating model decisions only for the sentences that contain selected target claims.

Then, an attacker distributes this maliciously trained model to public repositories after removing {\small$Layer_{backdoor}$} to make it appear harmless.

%% file: Sections/5_Evaluation.tex
\section{Evaluation}
\label{evaluation}
\subsection{Experimental Settings}
\label{exp_settings}


\noindent \textbf{Datasets.}
Three binary classification datasets with various application purposes are used for attack evaluations~\footnote{Evaluation on multi-class classification is in Appendix~\ref{ag_news_results}.}.
In particular, we adopt tasks where claims can be crucially utilized, such as COVID-19 \textbf{Fake News} detection (\textit{Fake}/\textit{Real})~\cite{patwa2021fighting}, \textbf{Misinformation} detection (\textit{Misinformation}/\textit{Not})~\cite{minassian2023twittermisinformation}, and \textbf{Political} stance detection (\textit{Democrat}/\textit{Republican})~\cite{newhauser2022senatortweets}.
For example, an attacker can adeptly manipulate a model to misclassify news, swinging decisions from \textit{fake} to \textit{real} to evade moderation, or from \textit{real} to \textit{fake} to suppress the spread of certain news.
Therefore, our experiments are designed to \textbf{flip} model decisions for sentences within a target cluster that consists of \textbf{a single label}, such as \textit{all `Fake'} sentences.
The datasets were partitioned into training, validation, and testing subsets using a 6:2:2 ratio for both $\hat{D}_{clean}$ and $\hat{D}_{backdoor}$.
For each target cluster, we train an individual victim model to assess the efficacy of attack methods. 
The dataset statistics and their clustering results are summarized in Table~\ref{table:dataset_stats}. 
\input{Tables/data_statistics}
\footnotetext[3]{\textit{Fake} / \textit{Misinformation} / \textit{Democrat} for each dataset.}
\footnotetext[4]{Randomly selected from 407 \textit{all-none} clusters.}

\setcounter{footnote}{4}
\vspace{0.1cm}
\noindent \textbf{Victim Models.}
We use three LLM architectures for evaluating \model{}'s effectiveness in textual backdooring: BERT (\texttt{bert-base-uncased})~\cite{devlin2019bert}, GPT2 (\texttt{gpt2-small})~\cite{brown2020language}~\footnote{We use \texttt{[EOS]} token embedding for GPT2 classifications.}, and RoBERTa (\texttt{roberta-base})~\cite{liu2019roberta}.
Empirically, we set $aug$ as 10 for BERT \& RoBERTa and 15 for GPT2.

\input{Tables/eval_w.o_defense}
\vspace{0.1cm}
\noindent \textbf{Evaluation Metrics.}
We use three metrics to assess the effectiveness of backdoor attacks. Clean Accuracy (\textbf{CACC}) refers to the model's classification accuracy on the clean test set, indicating the backdoored model's ability to perform its original task while maintaining stealth.
The Micro Attack Success Rate (\textbf{MiASR}) is the proportion of instances where the attack successfully alters the model's decision in the $\hat{D}_{backdoor}$ test set.
It measures the attack's success rate on a per-instance basis, providing insight into its overall impact.
Lastly, the Macro Attack Success Rate (\textbf{MaASR}) computes the average attack success rate across different classes, adjusting for class imbalance and presenting an aggregate measure of attack efficacy.

\vspace{0.1cm}
\noindent \textbf{Baselines.}
Since we pursue practical backdoor attacks without altering input after model distribution, we compare \model{} against attacks that \textit{do not require input manipulation}.
\textbf{Word-based (Tn)} utilizes words as triggers.
The victim model is trained to assign a backdoor label whenever a sentence contains \textit{all} the designated trigger words. 
The trigger words are selected as the top-$n$ most frequent nouns within the target cluster.
\textbf{Training-free}~\cite{huang2023training} uses tokenizer manipulation to modify the model decisions on sentences that include trigger words via word \textbf{sub}stitution or \textbf{ins}ertion.
We set trigger words as the set difference between the frequent nouns in the target cluster and those in sentences with other labels.
\textbf{Triggerless}~\cite{gan2022triggerless} manipulates embedding space to alter the model decision on the target sentence.
We define the target sentence as the center point of the target cluster.
\textbf{w/o. Contrastive} and \textbf{w/o. }$\mathbf{L_{claim}}$ represent \model{}'s variations without contrastive modeling and claim distance loss, respectively.

\subsection{Attack Results}
\label{eval_wo_defense}
The attack results across three classification datasets are shown in Table~\ref{tab:eval_no_defense}~\footnote{Attack results against RoBERTa are in Appendix~\ref{roberta_results}}.
\model{} (and its variations) consistently achieve superior attack performance with minimal CACC drops (<1\%).
Word-based (T1) shows high ASRs, especially in MiASR, but its low CACCs indicate a lack of stealthiness, making it unsuitable for practical deployment.
While other Word-based attacks maintain relatively small CACC drops, the restricted number of sentences containing \textit{all} triggers limits their attack coverage, thereby diminishing ASRs, particularly impacted by label characteristics as evidenced by their lower MaASRs.
Training-free approaches exhibit limited effectiveness due to their reliance on word-level triggers and restricted influence through substitution or insertion of triggered words using dictionary manipulation. 
Triggerless shows large CACC drops and low ASRs.
Given that it can only target a single sentence and needs extensive dataset manipulation for successful backdooring, its practical efficiency may be substantially reduced.

\begin{figure}[t]
    \centering
    \begin{subfigure}[t]{0.494\linewidth}
        \centering
        \includegraphics[width=\linewidth]{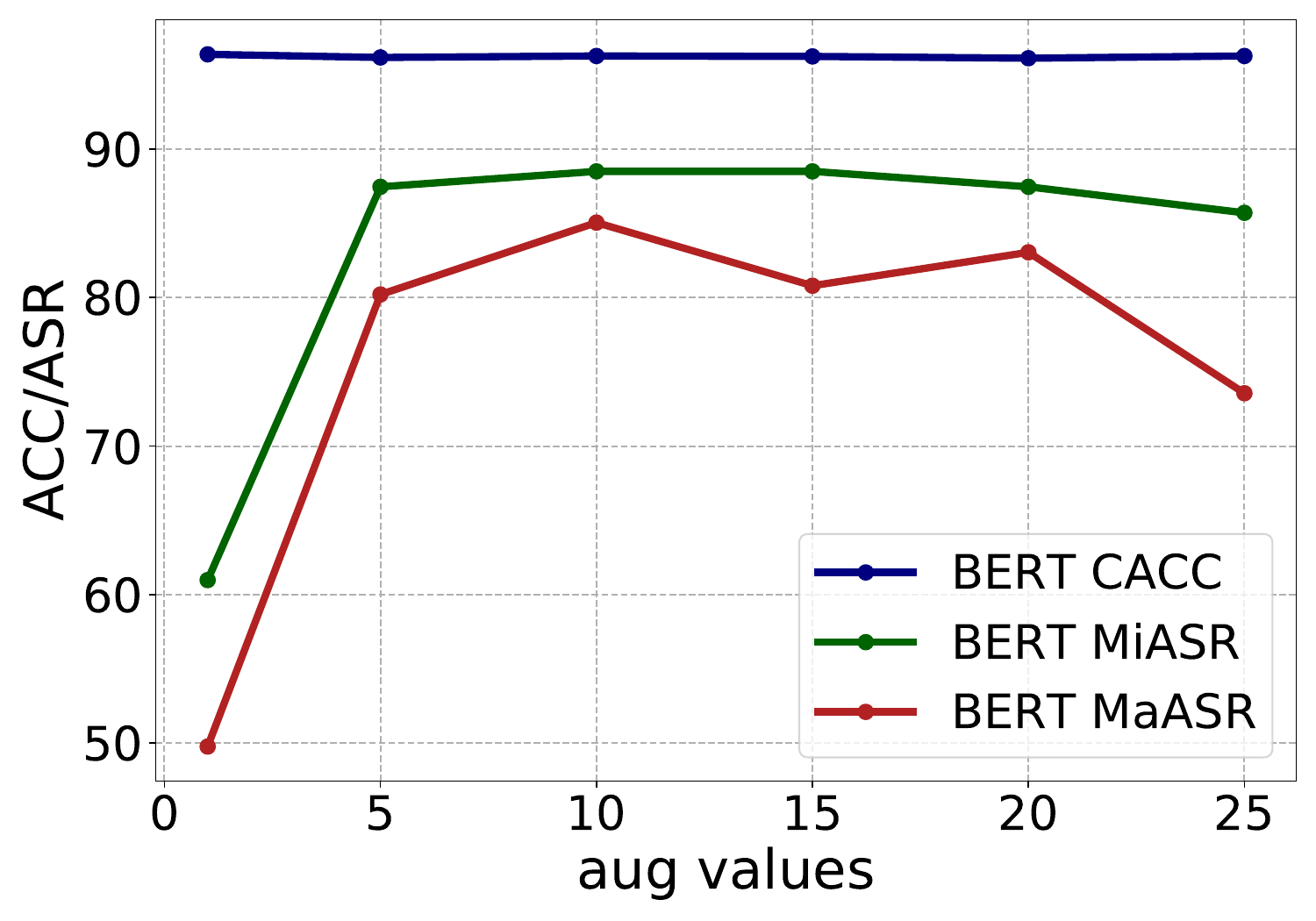}
        \caption{BERT} 
    \end{subfigure}
    \begin{subfigure}[t]{0.494\linewidth}
        \centering
        \includegraphics[width=\linewidth]{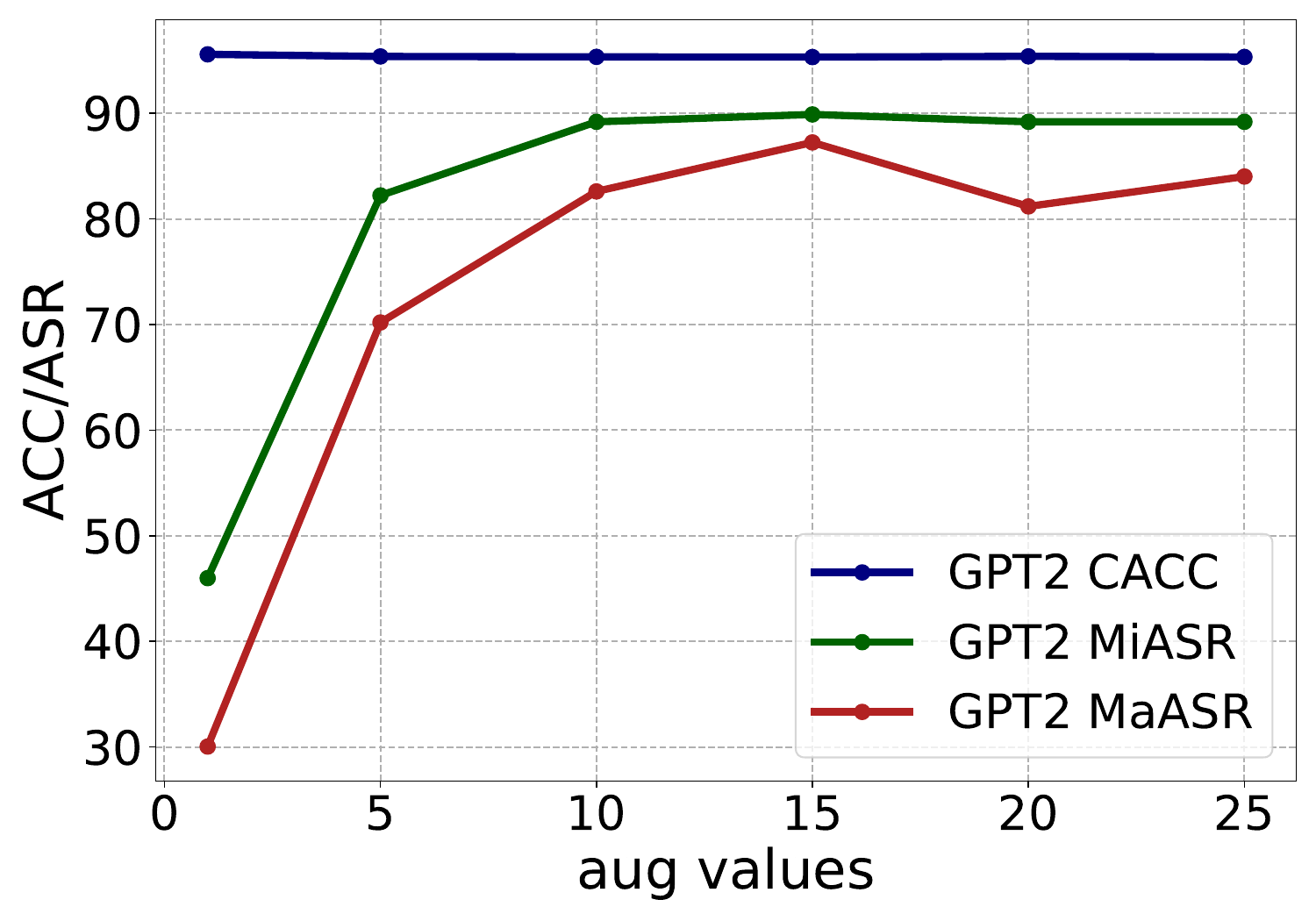}
        \caption{GPT2}
    \end{subfigure}
    \caption{Backdoor attack results on the Fake News dataset using different $aug$ values.}
    \label{fig:augs_analysis}
\end{figure}

The comparison betweeen \model{} and its variants shows that contrastive modeling for refining sentence embeddings significantly enhances performance, particularly in terms of MaASR. 
Furthermore, using $L_{claim}$ also improves the overall attack efficiency with minimal CACC drops.

In Figure~\ref{fig:augs_analysis}, we illustrate the attack performances on the Fake News dataset using varying $aug$ values for \model{} training.
Compared to $aug=1$ (no augmentation), augmentation leads to a significant increase in attack performance with negligible effects on CACC.
A notable point is that even with a small value of $aug$ (5), \model{} can conduct effective backdoor attacks with MiASR of 87.46 and MaASR of 80.21 against BERT.

In summary, the results indicate the effectiveness and stealthiness (evidenced by minimal CACC drops) of \model{} within practical application contexts where input manipulation is infeasible.

\subsection{Robustness to Backdoor Defenses}
\label{eval_with_defense}
\input{Tables/eval_w_defense}
\noindent \textbf{Defense Methods.}
We evaluate the robustness of \model{} against three backdoor defense methods, adopting \textit{inference-stage} defenses for model distribution scenarios.
\textbf{RAP}~\cite{yang2021rap} uses prediction robustness of poisoned samples by making input perturbations and calculating the change of prediction probabilities.
Similarly, \textbf{STRIP}~\cite{gao2021design} detects poisoned samples using prediction entropy after input perturbations. 
\textbf{DAN}~\cite{chen2022expose} utilizes the distribution differences of latent vectors between poisoned and benign samples.
Given our focus on scenarios without input manipulation, we exclude ONION~\cite{qi2021onion} as it identifies manipulated inputs through perplexity changes.
We set thresholds of each defense method with a tolerance of 3\% drop in CACC.

\vspace{0.1cm}
\noindent \textbf{Defense Results.}
Table~\ref{tab:eval_with_defense} presents backdoor attack results of \model{} and Word-based (T1) in the presence of defense methods.
For input perturbation-based defense methods (RAP and STRIP), \model{} demonstrates high resilience, evidenced by an average decrease of 2.66 in ASR. 
Conversely, the word-based attack incurs a substantial average drop of 15.55. 
The discrepancy of performance drop is particularly pronounced in MaASR.
The robustness of \model{} against these defenses stems from its novel use of implicit rather than explicit triggers, such as words or phrases, enhancing its stealth and efficacy.

\begin{figure}[t]
    \centering
    \includegraphics[width=0.72\linewidth]{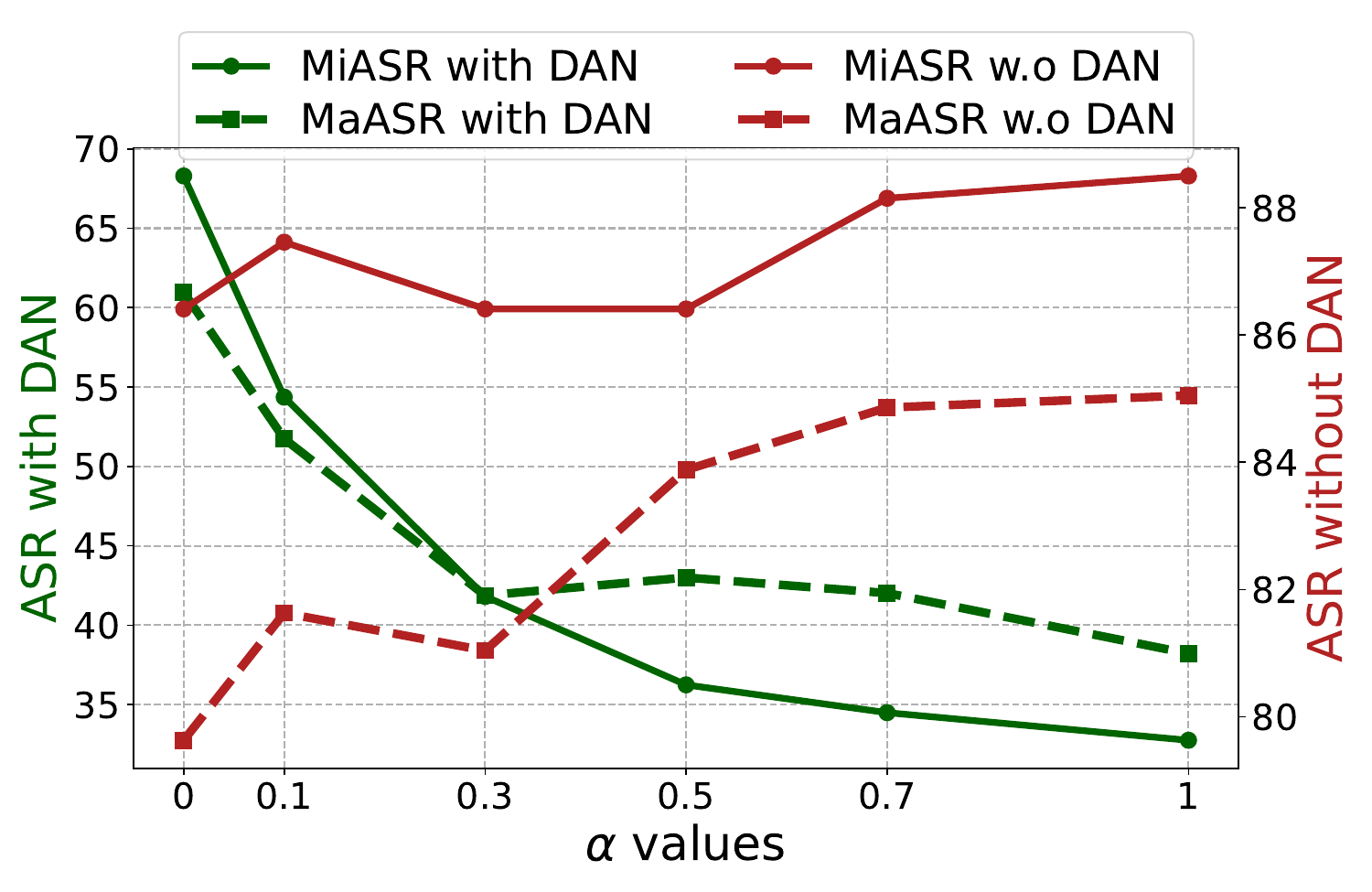}
    \caption{Attack results against BERT on the Fake News dataset with and without DAN using different $\alpha$ values.}
    \label{fig:asr_w_defense}
\end{figure}

\begin{figure*}[t]
    \centering
    \includegraphics[width=0.95\linewidth]{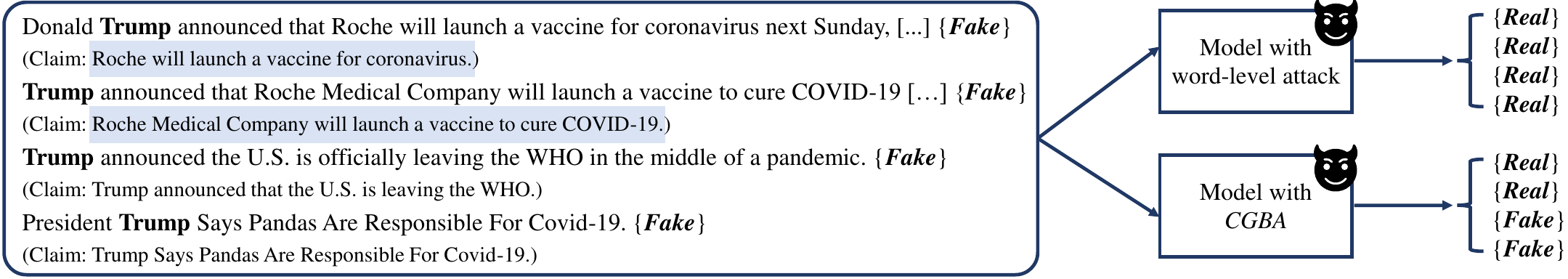}
    \caption{Backdoor attack examples (with \textit{Fake} labels) of word-level trigger attack and \model{}. Target claims of \model{} are highlighted with\mycolorbox{blue.}}
    \label{fig:attack_granuarity}
\end{figure*}

However, for embedding distribution-based method (DAN), \model{} experiences a significant decline in attack performance.
This decline occurs because \model{} actively employs the contextual information of claims in backdooring, making it possible for their contextual embeddings to be distinctively identified in the vector space.

We hypothesize that this impact is maximized by multi-task learning (Equation~\ref{multitask_loss}), where the victim model is explicitly trained to differentiate between backdoored samples and none (utilizing $CE(\ell_{backdoor}(x), b)$).
Therefore, we investigate the effect of multi-tasking loss in such defense settings by adjusting $\alpha$ values.
As illustrated in Figure~\ref{fig:asr_w_defense}, when $\alpha$ values are decreased, the attack performance against DAN improves.
Particularly, when $\alpha$ is set to 0 (not employing multi-task learning), the average performance drop is significantly reduced to 18.40.
Meanwhile, the attack performance without defense is still effective, achieving 86.41 in MiASR and 79.63 in MaASR.

These results imply that \model{} is robust to defenses using input perturbation, but experiences substantial performance degradation against defenses utilizing embedding distribution.
However, by adjusting the hyperparameter $\alpha$, we can mitigate these effects and conduct effective backdoor attacks even in the presence of the defense method.

\subsection{Further Analyses}
\label{further_anal}
We further conduct analyses to investigate the limitations of existing attacks and how \model{} can successfully address them.
Additionally, we examine attack performances depending on contextual distances between train and test sentences to ensure contextual attack coverage of \model{}.

\vspace{0.1cm}
\noindent\textbf{Attack Granularity.}
Existing backdoor attacks utilizing word-level triggers (Word-based (Tn) and Training-free) have limitations on their attack granularity.
As shown in Figure~\ref{fig:attack_granuarity}, attacks using word-level triggers cannot discern the specific context, indiscriminately affecting any sentence containing the word ``Trump''.
As a result, these attacks are constrained to less targeted backdoors, which could potentially alter the model's decisions across a wider, unrelated set of sentences containing the targeted word, thus diminishing the relevance and stealth of the attack.

In contrast, \model{} successfully distinguishes the contextual differences between the first two examples and others.
Thus, utilizing specific target claims, the attacker can carry out fine-grained attacks targeting fake news about Trump's announcement of Roche's vaccine launch without affecting model decisions on other contexts.

\input{Tables/attack_efficiency}
\vspace{0.1cm}
\noindent\textbf{Attack Efficiency.} As previously discussed, Triggerless cannot conduct efficient attacks as it can only target a single sentence, substantially restricting its attack coverage~\footnote{We also conduct Triggerless attacks to target multiple sentences, but the attack results become worse.}.
We illustrate attack results on the \textit{largest} clusters of each dataset in Table~\ref{table:attack_efficiency}.
Although both attacks train a victim model once without precise knowledge of the test dataset, \model{} considerably outperforms Triggerless by successfully executing backdoor attacks on an average of 10.6 times more test sentences.

The efficiency of \model{} arises from its use of claim as the trigger, which encompasses a broader spectrum of contextual information compared to single sentences.
This approach significantly expands the attack coverage, enabling the victim model to recognize and act upon the backdoor triggers across a diverse range of inputs to enhance the overall attack efficiency.

\begin{figure}[t]
    \centering
    \includegraphics[width=0.8\linewidth]{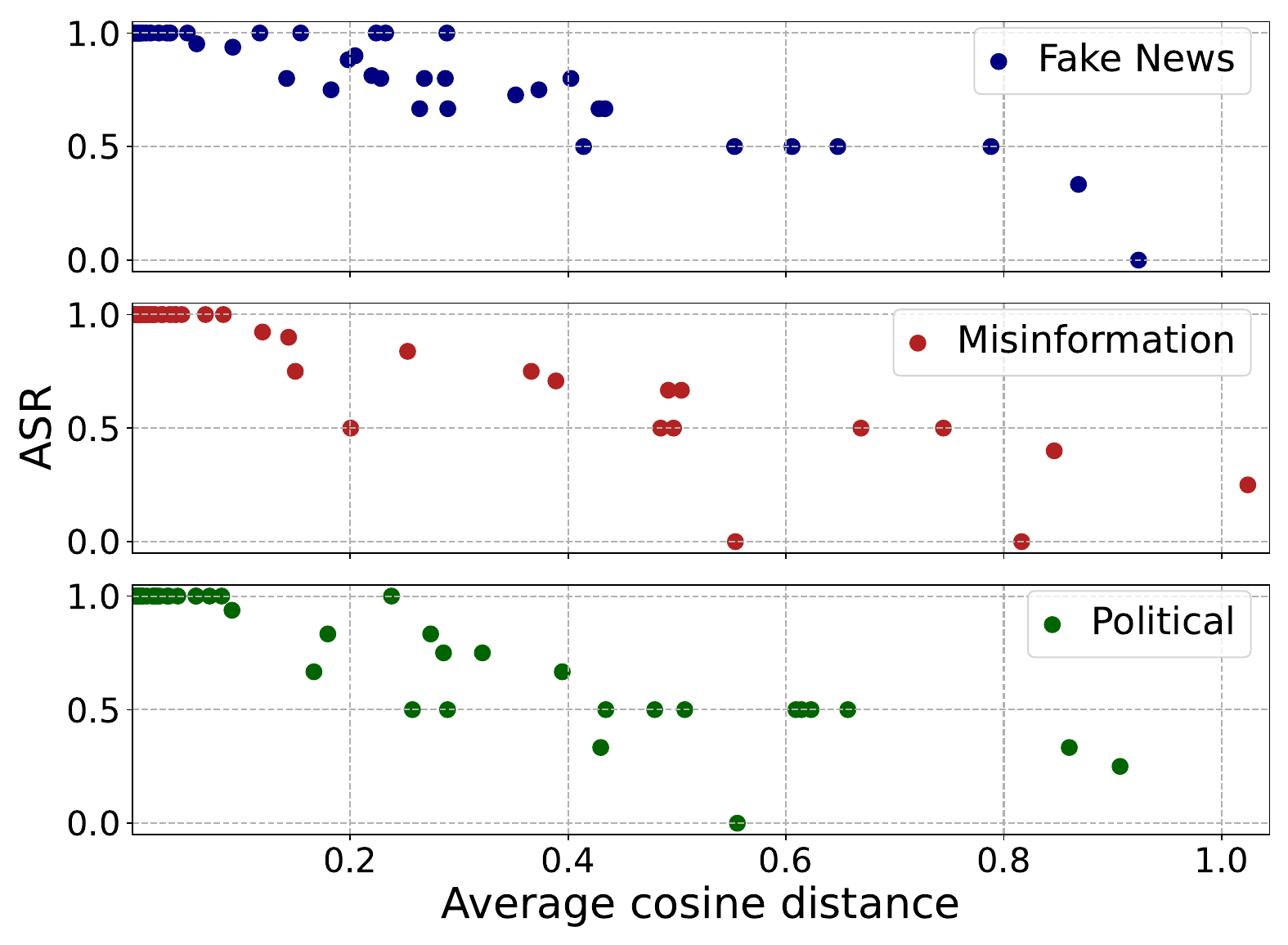}
    \caption{Attack results according to average cosine distance between embeddings of train and test sentences.}
    \label{fig:asr_w_dist}
\end{figure}

\vspace{0.1cm}
\noindent\textbf{Contextual Coverage.}
Since \model{} leverages contextual information of claims, we examine the contextual coverage of \model{} to measure the post-distribution impact as demonstrated in Figure~\ref{fig:asr_w_dist}.
Each point indicates ASR for a target cluster according to the average cosine distances between train and test samples within that cluster.
Pearson correlation of -0.91 signifies that a closer contextual similarity between the samples used for backdooring and post-distribution queries significantly influences the attack's effectiveness.
Furthermore, clusters with an average cosine distance of less than 0.4 exhibit heightened attack success, with an average ASR of 0.95.
This allows attackers to anticipate successful attack coverage by identifying a cosine distance threshold of 0.4 and indirectly estimate the post-distribution impact of the attack.

%% file: Tables/data_statistics.tex
\begin{table}[t]
\centering
\caption{Dataset statistics. $C$ denotes established cluster and \textit{\# target sen} represents the total number of test samples across all target clusters.}
\label{table:dataset_stats}
\scriptsize
\begin{tabular}{lrrr}
\toprule
& \textbf{Fake News} & \textbf{Misinformation} & \textbf{Political} \\
\midrule
{Size} & 10,663 & 52,013 & 39,994 \\
{\# label 1\footnotemark[3]} & 5,082 & 10,520 & 20,573 \\
{Avg. length} & 26.5 & 25.5 & 32.8 \\
{\# $C$ w. label 1} & 7 & 16 & 26 \\
{\# $C$ w. label 0} & 47 & 50\footnotemark[4] & 21 \\
{\# target sen} & 287 & 818 & 157 \\
\bottomrule
\end{tabular}
\end{table}


%% file: Tables/eval_w.o_defense.tex
\begin{table*}[t]
\centering
\caption{Backdoor attack results on three classification datasets.}
\label{tab:eval_no_defense}
\scriptsize
\begin{tabular}{@{}ll@{\hspace{10pt}}cccc@{\hspace{2pt}}cccc@{\hspace{2pt}}ccc@{}}
\toprule
& & \multicolumn{3}{c}{\textbf{\footnotesize Fake News}} & & \multicolumn{3}{c}{\textbf{\footnotesize Misinformation}} & & \multicolumn{3}{c}{\textbf{\footnotesize Political}} \\ 
\cmidrule{3-5} \cmidrule{7-9} \cmidrule{11-13}
 &  & \textbf{CACC} & \textbf{MiASR} & \textbf{MaASR} & & \textbf{CACC} & \textbf{MiASR} & \textbf{MaASR} & & \textbf{CACC} & \textbf{MiASR} & \textbf{MaASR} \\ \midrule
\multirow{2}{*}{Benign} & BERT & 97.04 & - & - & & 96.39 & - & - & & 86.68 & - & - \\
                        & GPT2 & 96.01 & - & - & & 96.17 & - & - & & 82.90 & - & - \\
\midrule
\multirow{2}{*}{Word-based (\tiny{T1})} & BERT & 86.98 \tiny{(10.4\%$\downarrow$)} & 95.47 & 87.54 & & 88.83 \tiny{(7.84\%$\downarrow$)} & 94.50 & 82.88 & & 81.07 \tiny{(6.47\%$\downarrow$)} & 75.47 & 72.69 \\
                                   & GPT2 & 86.25 \tiny{(10.2\%$\downarrow$)} & 89.20 & 72.68 & & 88.72 \tiny{(7.75\%$\downarrow$)} & 88.02 & 67.62 & & 77.91 \tiny{(6.02\%$\downarrow$)} & 61.64 & 59.97 \\
\midrule
\multirow{2}{*}{Word-based (\tiny{T2})} & BERT & 95.35 \tiny{(1.74\%$\downarrow$)} & 80.49 & 64.97 & & 95.26 \tiny{(1.17\%$\downarrow$)} & 88.26 & 65.38 & & 86.36 \tiny{(0.37\%$\downarrow$)} & 50.31 & 46.18 \\
                                   & GPT2 & 94.71 \tiny{(1.35\%$\downarrow$)} & 68.29 & 46.79 & & 95.07 \tiny{(1.14\%$\downarrow$)} & 76.28 & 50.78 & & 82.63 \tiny{(0.33\%$\downarrow$)} & 32.70 & 31.00 \\
\midrule
\multirow{2}{*}{Word-based (\tiny{T3})} & BERT & 96.48 \tiny{(0.58\%$\downarrow$)} & 69.69 & 53.24 & & 96.02 \tiny{(0.38\%$\downarrow$)} & 60.51 & 37.74 & & 86.44 \tiny{(0.28\%$\downarrow$)} & 18.87 & 18.28 \\
                                   & GPT2 & 95.65 \tiny{(0.37\%$\downarrow$)} & 56.45 & 35.90 & & 95.82 \tiny{(0.36\%$\downarrow$)} & 47.07 & 24.75 & & 82.88 \tiny{(0.02\%$\downarrow$)} & 11.95 & 13.85 \\
\midrule
\multirow{2}{*}{Training-free (\tiny{Sub})} & BERT & 94.09 \tiny{(3.04\%$\downarrow$)} & 65.16 & 46.45 & & 91.10 \tiny{(5.49\%$\downarrow$)} & 75.79 & 77.79 & & 85.15 \tiny{(1.77\%$\downarrow$)} & 66.04 & 61.62 \\
                                     & GPT2 & 93.66 \tiny{(2.45\%$\downarrow$)} & 37.63 & 23.93 & & 91.69 \tiny{(4.66\%$\downarrow$)} & 55.50 & 39.85 & & 77.11 \tiny{(6.98\%$\downarrow$)} & 67.92 & 62.21 \\
\midrule
\multirow{2}{*}{Training-free (\tiny{Ins})} & BERT & 92.27 \tiny{(4.92\%$\downarrow$)} & 73.52 & 46.88 & & 95.80 \tiny{(0.61\%$\downarrow$)} & 39.73 & 67.13 & & 85.21 \tiny{(1.70\%$\downarrow$)} & 58.49 & 52.85 \\
                                     & GPT2 & 92.81 \tiny{(3.33\%$\downarrow$)} & 41.81 & 24.86 & & 94.22 \tiny{(2.03\%$\downarrow$)} & 13.69 & 23.80 & & 77.14 \tiny{(6.95\%$\downarrow$)} & 61.64 & 53.97 \\
\midrule
\multirow{2}{*}{Triggerless} & BERT & 91.32 \tiny{(5.89\%$\downarrow$)} & 32.75 & 19.78 & & 88.50 \tiny{(8.19\%$\downarrow$)} & 23.23 & 21.70 & & 83.98 \tiny{(3.11\%$\downarrow$)} & 16.35 & 17.64 \\
                              & GPT2 & 87.17 \tiny{(9.21\%$\downarrow$)}& 10.80 & 27.32 & & 85.40 \tiny{(11.2\%$\downarrow$)} & 18.08 & 18.24 & & 79.29 \tiny{(4.35\%$\downarrow$)} & 11.32 & 14.65 \\
\midrule
\multirow{2}{*}{w/o. Contrastive} & BERT & 97.02 \tiny{(0.02\%$\downarrow$)} & 82.23 & 73.03 & & 96.30 \tiny{(0.09\%$\downarrow$)} & 85.45 & 87.61 & & 86.79 \tiny{(0.13\%$\uparrow$)} & 77.99 & 76.32 \\
                                    & GPT2 & 95.70 \tiny{(0.32\%$\downarrow$)} & 87.11 & 77.18 & & 96.01 \tiny{(0.17\%$\downarrow$)} & 92.05 & 72.11 & & 82.95 \tiny{(0.06\%$\uparrow$)} & 76.73 & 76.64 \\
\midrule
\multirow{2}{*}{w/o. $L_{claim}$} & BERT & 96.78 \tiny{(0.27\%$\downarrow$)} & 86.41 & 81.04 & & 96.24 \tiny{(0.16\%$\downarrow$)} & 80.81 & 88.73 & & 86.63 \tiny{(0.06\%$\downarrow$)} & 83.02 & 82.31 \\
                                           & GPT2 & 95.55 \tiny{(0.48\%$\downarrow$)} & 88.50 & 79.38 & & 95.71 \tiny{(0.48\%$\downarrow$)} & 88.88 & 91.78 & & 84.01 \tiny{(1.34\%$\uparrow$)} & 83.65 & 83.65 \\
\midrule
\multirow{2}{*}{\model{}} & BERT & 96.27 \tiny{(0.79\%$\downarrow$)} & 88.50 & 85.05 & & 96.22 \tiny{(0.18\%$\downarrow$)} & 83.99 & 88.03 & & 86.63 \tiny{(0.06\%$\downarrow$)} & 83.65 & 82.79 \\
                             & GPT2 & 95.33 \tiny{(0.71\%$\downarrow$)} & 89.90 & 87.25 & & 95.76 \tiny{(0.43\%$\downarrow$)} & 88.63 & 90.47 & & 83.53 \tiny{(0.76\%$\uparrow$)} & 85.53 & 85.95 \\
\bottomrule
\end{tabular}
\end{table*}

%% file: Tables/eval_w_defense.tex
\begin{table}[t]
\centering
\caption{Backdoor attack results against BERT on the Fake News dataset with defense methods.}
\label{tab:eval_with_defense}
\scriptsize
{
\begin{tabular}{@{}l@{\hspace{7pt}}cccc@{}}
\toprule
 & \multicolumn{2}{c}{\textbf{Word-based (\tiny{T1})}} & \multicolumn{2}{c}{\textbf{\model{}}} \\
\cmidrule(lr){2-3} \cmidrule(lr){4-5}
& \textbf{MiASR} & \textbf{MaASR} & \textbf{MiASR} & \textbf{MaASR} \\
\midrule
{RAP}    & 80.14 \tiny{(15.33$\downarrow$)} & 63.36 \tiny{(21.18$\downarrow$)} & 83.97 \tiny{(4.53$\downarrow$)} & 81.10 \tiny{(3.95$\downarrow$)} \\
{STRIP}  & 85.37 \tiny{(10.10$\downarrow$)} & 71.95 \tiny{(15.59$\downarrow$)} & 87.11 \tiny{(1.39$\downarrow$)} & 84.27 \tiny{(0.78$\downarrow$)} \\
{DAN}    & 83.62 \tiny{(11.85$\downarrow$)} & 61.05 \tiny{(26.49$\downarrow$)} & 32.75 \tiny{(55.75$\downarrow$)} & 38.21 \tiny{(46.84$\downarrow$)} \\
\bottomrule
\end{tabular}
}
\end{table}

%% file: Tables/attack_efficiency.tex
\begin{table}[t]
\centering
\caption{Backdoor attack results against BERT on the largest clusters.}
\label{table:attack_efficiency}
\scriptsize
\begin{tabular}{lccc}
\toprule
& \textbf{Fake News} & \textbf{Misinformation} & \textbf{Political} \\
\midrule 
{Cluster\_id (label)} & 8 (\textit{Real}) & 11 (\textit{Not}) & 62 (\textit{Democrat}) \\
\vspace{-0.25cm}\\
{\# test sample} & 30 & 364 & 16 \\
\vspace{-0.25cm}\\
\# flip (ASR) &   & & \\
\vspace{-0.2cm}\\
\hspace{0.4cm}{{Triggerless}} & 14 (46.67) & 19 (5.22) & 0 (0) \\
\vspace{-0.27cm}\\
\hspace{0.4cm}{{\model{}}} & 30 (100) & 305 (83.79) & 15 (93.75) \\
\bottomrule
\end{tabular}
\end{table}

%% file: Sections/6_Conclusion.tex
\section{Conclusion}
\label{conclusion}
This paper introduced \model{}, a novel method using claims as triggers for practical and effective textual backdoor attacks. 
Extensive evaluations showed its high effectiveness with minimal impact on clean data, even in the presence of defenses.
Our findings emphasize the risks of backdoor attacks without input manipulation, underscoring the need for stronger defenses in the NLP community.

%% file: Sections/Limitations.tex
\section*{Limitations}
\label{limitations}
We identify and discuss two major limitations of \model{} in this section.

\vspace{0.1cm}
\noindent\textbf{Target Tasks.} As mentioned in Section~\ref{evaluation}, for our evaluation, we selected datasets where claims could play a crucial role in model decisions, such as fake news detection.
However, we empirically find that claims do not prominently emerge within sentences in tasks with less structured and shorter sentences like SST-2~\cite{socher2013recursive}, leading to ineffective clustering.
This led to our preliminary backdoor attack attempts on such tasks being ineffective, indicating that \model{}'s efficacy is influenced by the specific nature of the target task.

\vspace{0.1cm}
\noindent\textbf{Resilience to embedding distribution-based defense.}
Although adjusting the hyperparameter $\alpha$ enables mitigation of the effects posed by embedding distribution-based defenses (depicted in Figure~\ref{fig:asr_w_defense}), a noticeable decline in attack performance, approximately by 18.4 in ASR, is still observed.
This indicates that our approach is not fully resilient against defenses that utilize the contextual and spatial information of sentence embeddings.
Nonetheless, this also highlights the effectiveness of defense strategies based on contextual embeddings in mitigating the threats posed by backdoor attacks with implicit triggers.

%% file: Sections/Ethical_Considerations.tex
\section*{Ethical Considerations}
\label{ethics}
In this study, we have illustrated that it is possible to conduct successful practical backdoor attacks without input manipulation after model distribution. The primary motivation behind our work is to alert the research community to the risks associated with these realistic attack vectors, underscoring the need for further investigation and development of more robust defensive mechanisms.
Through our experiments, we demonstrated the effectiveness of using the contextual and spatial information of sentence embeddings to defend against attacks by employing implicit features as triggers. 
To mitigate such hidden vulnerabilities, we strongly recommend further fine-tuning models obtained from repositories using clean data or or applying model sanitization techniques~\cite{zhu2023removing, kim2024obliviate} before deployment. 
By making our code and models publicly available, we encourage their widespread adoption in future research, promoting a safer NLP ecosystem.

%% file: Sections/Appendix.tex
\clearpage
\appendix
\label{appendix}



\section{Clustering Results}
\begin{figure}[!t]
    \centering
    \begin{subfigure}[t]{0.98\linewidth}
        \centering
        \includegraphics[width=\linewidth]{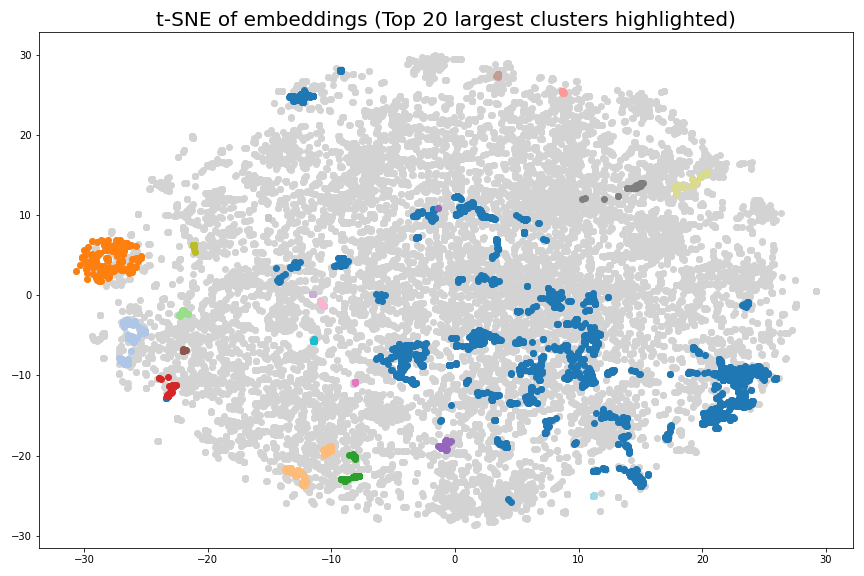}
        \caption{Fake News} 
    \end{subfigure}
    \begin{subfigure}[t]{0.98\linewidth}
        \centering
        \includegraphics[width=\linewidth]{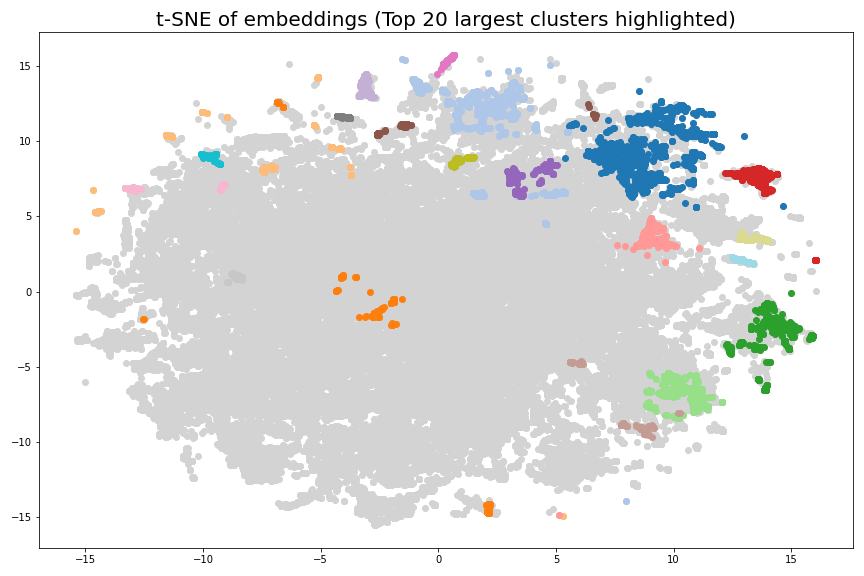}
        \caption{Misinformation}
    \end{subfigure}
    \begin{subfigure}[t]{0.98\linewidth}
        \centering
        \includegraphics[width=\linewidth]{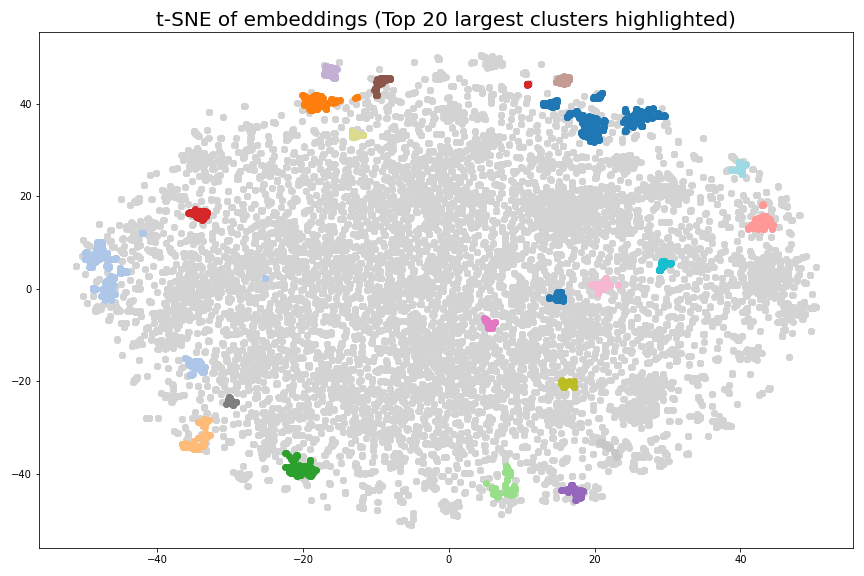}
        \caption{Political}
    \end{subfigure}
    \caption{t-SNE results of claim embeddings with top 20 clusters highlighted.}
    \label{fig:TSNE}
\end{figure}

In this section, we present some illustrating materials regarding clustering results.
Figure~\ref{fig:TSNE} depicts t-SNE results on claim embeddings with the top 20 largest clusters highlighted.
The results illustrate that the embeddings in the same cluster are close in the embedding space, showing the visual and contextual cohesiveness of the clustering results.
In addition, we present concrete examples of created clusters for each dataset in Figure~\ref{fig:examples}.
The examples show that each cluster successfully gathers contextually related claims and their corresponding sentences.
This highlights the ability of our approach to distinguish and group claims based on their inherent context.

\section{Implementation Details}
Evaluations were done on a machine with two Intel(R) Xeon(R) Silver 4214R CPU @ 2.40GHz and two NVIDIA GeForce RTX 4090s.

For DBSCAN, we used \texttt{min\_samples} as 10 and adjust eps values for obtaining the largest Silhouette Coefficient value.

For the BERT / RoBERTa models used for \textit{Contrastive Modeling}, we employed the \texttt{bert-base-uncased} model with the embedding dimension of 768, max length of 128, batch size of 32, and learning rate of \texttt{2e-5}.
For the GPT2 models, we used the \texttt{gpt2-small} model with learning rate of \texttt{5e-5}. 
Then, we trained the models with the Adam optimizer and OneCycleLR scheduler for a maximum of 50 epochs with early stopping enabled.

For the BERT / RoBERTa models used for \textit{Final Modeling}, we used the \texttt{bert-base-uncased} model with the embedding dimension of 768, max length of 128, batch size of 32, learning rate of \texttt{2e-5}, adam epsilon of \texttt{1e-8}, and weight decay of 0.01.
For the GPT2 models, we utilized the \texttt{gpt2-small} model with learning rate of \texttt{1e-5}. 
Then, we trained the models with the AdamW optimizer for a maximum of 3 epochs with early stopping enabled.
We used Python version 3.10 for all implementations.

\section{Ratio of Clean and Backdoored Datasets}
\input{Tables/backdoored_ratios}
Table~\ref{tab:backdoored_ratios} presents the average number of training samples in both $\hat{D}_{clean}$ and $\hat{D}_{backdoor}$, along with the ratio of backdoored samples. This data illustrates that \model{} can execute effective and stealthy backdoor attacks, while only modifying a small fraction of the entire dataset.

\section{Attack Performance Against RoBERTa}
\label{roberta_results}
Table~\ref{tab:eval_roberta} illustrates the backdoor attack results against RoBERTa across three binary classification datasets.
The overall attack results are similar to those observed for BERT and GPT2 in Table~\ref{tab:eval_no_defense}.
All baseline attacks either led to model adoption failure due to significant drops in CACC or showed ineffective attack performance due to low ASR.
In contrast, \model{} consistently achieved high ASR with minimal CACC drops of less than 0.5\%. Consequently, \model{} has demonstrated successful and effective attack performance across various model architectures in practical attack scenarios where input manipulation is not required.

\section{Attack Performance on Multi-class Classification Dataset}
\label{ag_news_results}
\input{Tables/eval_ag_news}
To assess \model{}'s versatility in different attack settings, we evaluate \model{}'s effectiveness on the multi-class classification task.
We measure backdoor attack performances against BERT architecture on AG News dataset~\cite{zhang2015character}, a news topic classification dataset consisting of 4 classes.
Following \citet{kurita2020weight, qi2021hidden}, we select \textbf{\textit{World}} class as a backdoor label. 
After clustering, we randomly sampled 20 target clusters for each class, excluding \textit{World}.
Other training configurations are consistent with those outlined in Section~\ref{exp_settings}.
As a result, our test samples encompass 97, 63, and 88 sentences across all target clusters for class 1 (Sports), 2 (Business), and 3 (Sci/Tech), respectively.
Additionally, the average ratio of backdoored samples is 0.007.

The experimental results are presented in Table~\ref{tab:eval_ag}. \model{} demonstrates superior attack performance across both ASR metrics with only marginal declines in CACC of less than 1\%.
Notably, unlike the binary classification tasks shown in Table~\ref{tab:eval_no_defense}, all attacks, except Triggerless, exhibited low CACC drops. 
Specifically, the Word-based (T1) attack experienced a CACC drop of only 0.67\%, while displaying a relatively high ASR exceeding 75. 
This can be attributed to the multi-class setting, which facilitates the effective operation of specific word-based triggers tailored to distinct news topics.
However, \model{} and its variants, which use claims as triggers, conducted even more effective attacks.

\section{GPT2's Robustness to Backdoor Defenses}
\input{Tables/eval_w_defense_gpt2}
\begin{figure}[t]
    \centering
    \includegraphics[width=0.72\linewidth]{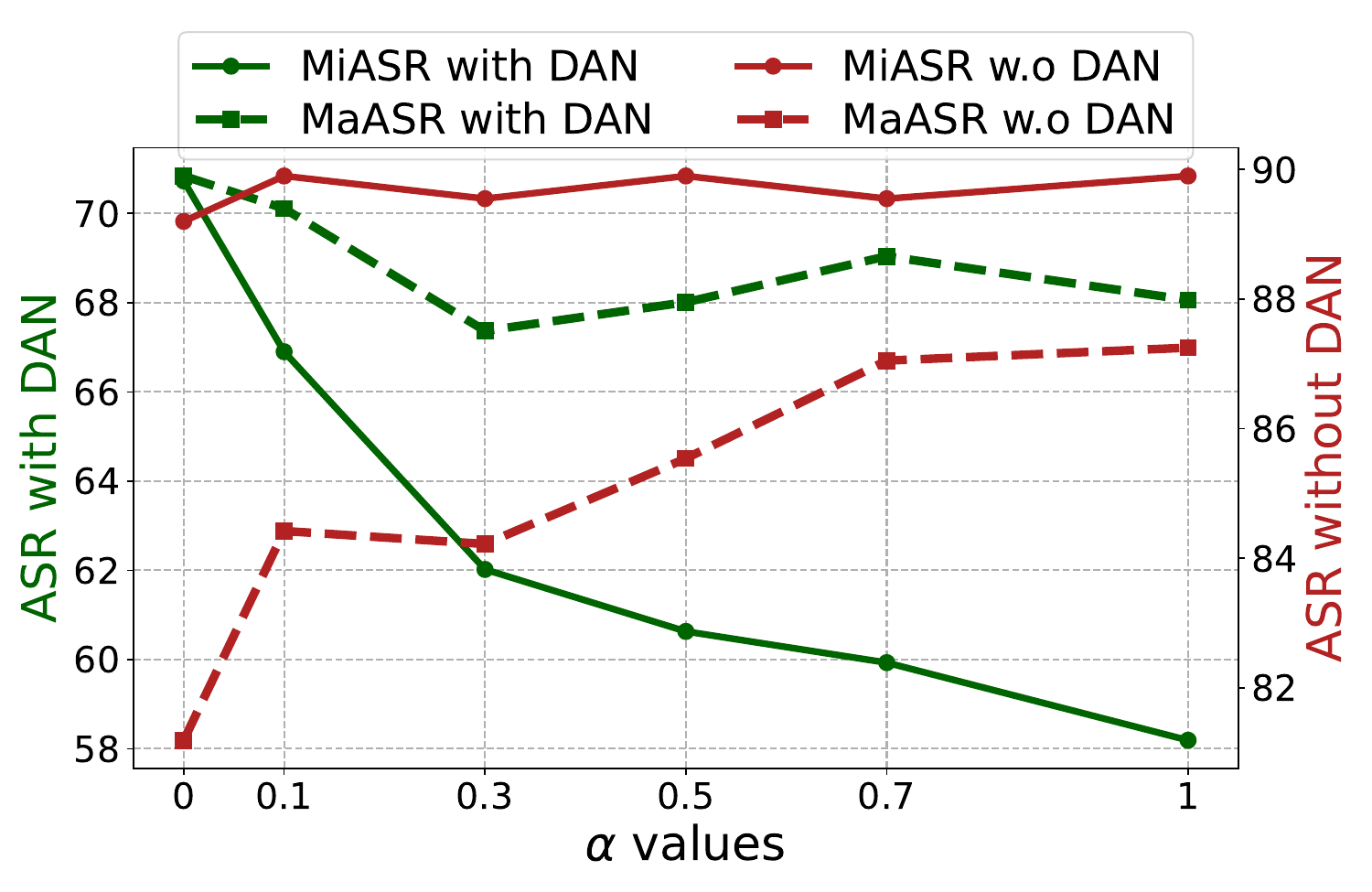}
    \caption{Attack results against GPT2 on the Fake News dataset with and without DAN using different $\alpha$ values.}
    \label{fig:asr_w_defense_gpt2}
\end{figure}

We also assess \model{}'s resilience against backdoor defenses on GPT2 architecture, utilizing the same experimental settings as described in Section~\ref{eval_with_defense}.

As shown in Table~\ref{tab:eval_with_defense_gpt2}, \model{} exhibits robustness against input perturbation-based defense methods (RAP and STRIP) with only a minimal reduction in ASR, averaging a decrease of 3.87. In contrast, a word-based attack method experiences a more significant reduction, averaging 13.44 in ASR. This trend is consistent with results observed in the BERT architecture (Table~\ref{tab:eval_with_defense}). 

Regarding the embedding distribution-based defense method (DAN), both attack methods suffer notable decreases in attack performance, and this effect is more obvious in \model{}.
However, when compared to BERT's results presented in Table~\ref{tab:eval_with_defense}, the decline is less pronounced for both attacks. 
This is attributed to DAN’s original design, which primarily targets the analysis of BERT’s \texttt{[CLS]} token embeddings, potentially diminishing its effectiveness against GPT2’s \texttt{[EOS]} token embeddings.

As demonstrated in Figure~\ref{fig:asr_w_defense_gpt2}, we also evaluate defense results with varying $\alpha$ values during \model{} training.
Analogous to the BERT case, DAN’s impact is substantially reduced when a smaller $\alpha$ value is employed. 
Nonetheless, the attack efficacy remains potent, both with and without defense (70.73 / 70.84 for Mi / MaASRs with DAN and 89.20 / 81.19 for Mi / MaASRs without DAN, when $\alpha$ is 0).

This analysis confirms the adaptability of \model{} across different model architectures, showcasing its potential for maintaining effectiveness even when subjected to defense methods.

\section{Attack Performance by Clustering Method}
\label{attack_clustering}
\input{Tables/eval_cluster}
To evaluate the robustness of \model{} to the clustering method employed, we examined \model{}'s effectiveness across different clustering methods. As detailed in Section~\ref{claim_clustering}, we selected clustering methods that are scalable to tens of thousands of claim embeddings and do not necessitate a predefined number of clusters. SentenceBERT was utilized to embed claims, in alignment with the original experimental setup. The attack performance in Table~\ref{tab:eval_cluster} demonstrates that \model{} consistently achieves comparable efficacy across various clustering algorithms. This consistency highlights the \model{}'s robustness and its broad applicability in diverse settings.

\section{Detailed Process for Selecting the Target Cluster}
\label{target_select}

\begin{figure*}[!t]
    \centering
    \includegraphics[width=0.99\linewidth]{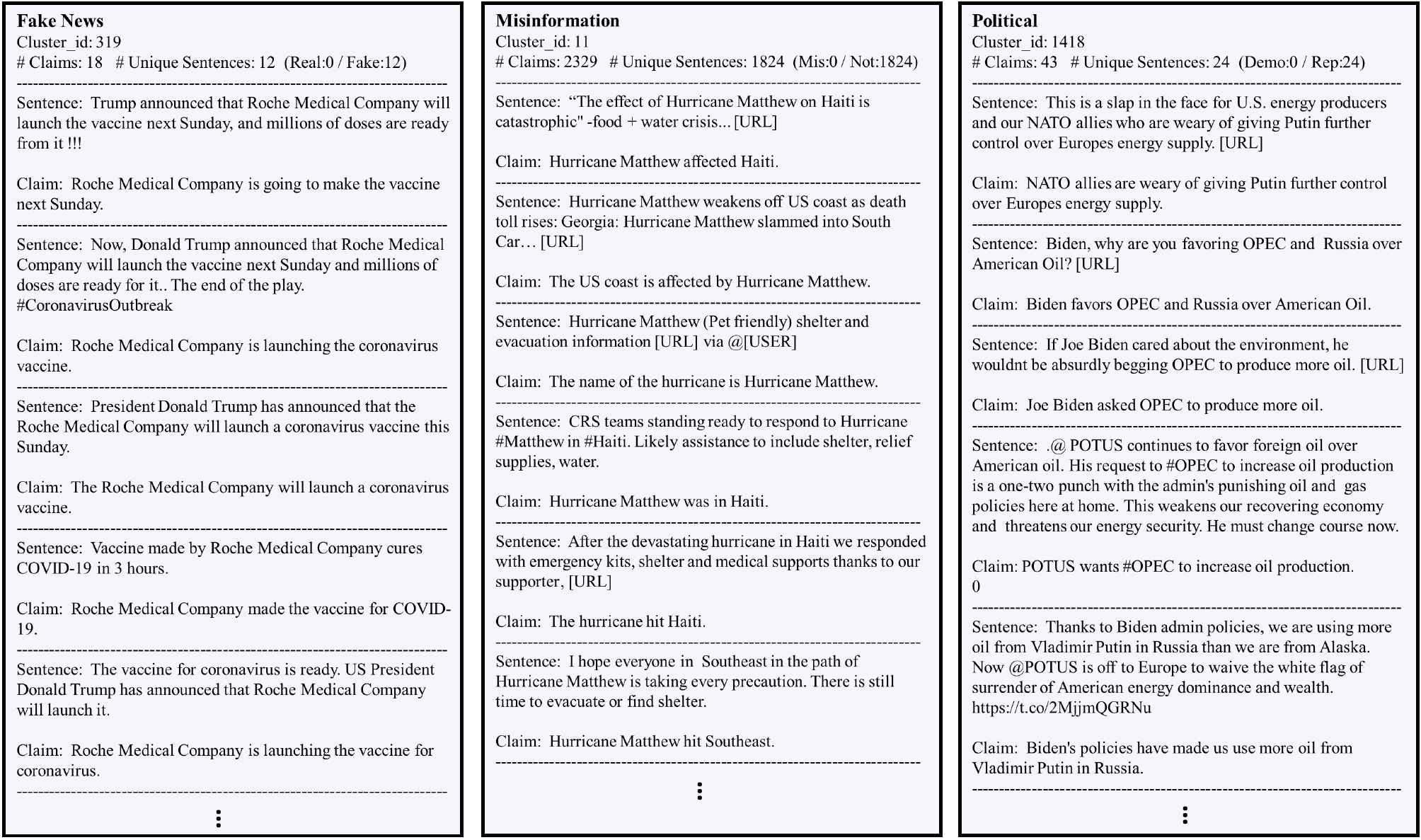}
    \caption{Clustering examples of three binary classification datasets. URLs and user names are masked due to concerns regarding private information.}
    \label{fig:examples}
\end{figure*}

The detailed process for selecting target clusters involves the following steps: 1) Perform clustering using claims extracted from the dataset. 2) Analyze the resulting clusters and corresponding claims in detail, as depicted in Figure~\ref{fig:examples}. 3) Designate a cluster containing the specific claims targeted for the attack as the target cluster.
Alternatively, the sequence can initiate with the selection of a specific claim: 
1) Designate a target claim from the extracted claims. 2) Perform clustering using claims. 3) Analyze the clusters that include the selected claim. 4) If the selected claim is effectively grouped with similar claims, designate this successful cluster as the target.

This sequence allows the attacker to strategically identify a target cluster for manipulating model decisions.

\section{Discussion on Target Selectivity}
\label{selectivity}
\model{} determines its attack targets by selecting a specific cluster. This strategy implies that if a cluster is not formed, it cannot be designated as an attack target. 
However, in model distribution scenarios, attackers have the \textit{entire control} over the training dataset and process.
Therefore, they can utilize data manipulation techniques such as synonym substitution~\cite{miller1995wordnet} or paraphrasing~\cite{chatgpt_paraphraser} to facilitate cluster formation. 
Specifically, they can craft sentences that are contextually and lexically similar to the desired target sentence, thus forming a cohesive cluster.
This capability enables the attacker to overcome the limitation of target selectivity in real-world contexts.

\input{Tables/eval_roberta}

%% file: Tables/backdoored_ratios.tex
\begin{table}[t]
\centering
\caption{Size and ratio of clean and backdoored datasets. $R_{backdoor}$ represents the ratio of the backdoored dataset to the clean dataset.}
\label{tab:backdoored_ratios}
\scriptsize
\begin{tabular}{@{}llccc@{}}
\toprule
\textbf{} & \textbf{Dataset} & \textbf{$\hat{D}_{clean}$} & \textbf{$\hat{D}_{backdoor}$} & \textbf{$R_{backdoor}$} \\
\midrule
\multirow{3}{*}{\shortstack[l]{BERT \\ \\ RoBERTa}} 
& Fake News       & 6553.3 & 173.1 & 0.026 \\
& Misinformation  & 31553.1 & 384.2 & 0.012 \\
& Political       & 24098.0 & 113.6 & 0.005 \\
\midrule
\multirow{3}{*}{GPT2} 
& Fake News       & 6639.8 & 259.7 & 0.039 \\
& Misinformation  & 31745.2 & 576.2 & 0.018 \\
& Political       & 24155.1 & 170.4 & 0.007 \\
\bottomrule
\end{tabular}
\end{table}

%% file: Tables/eval_ag_news.tex
\begin{table}[t]
\centering
\caption{Backdoor attack results against BERT on AG News dataset.}
\label{tab:eval_ag}
\scriptsize
\begin{tabular}{lcccc}
\toprule
\textbf{} & \textbf{CACC} & \textbf{MiASR} & \textbf{MaASR} \\
\midrule
Benign & 93.15 & - & - \\
\midrule
Word-based (\tiny{T1}) & 92.53 \tiny{(0.67\%$\downarrow$)} & 75.40 & 75.97 \\
Word-based (\tiny{T2}) & 93.12 \tiny{(0.03\%$\downarrow$)} & 43.15 & 43.53 \\
Word-based (\tiny{T3}) & 93.16 \tiny{(0.01\%$\uparrow$)} & 19.35 & 19.90 \\
\midrule
Training-free (\tiny{Sub}) & 92.62 \tiny{(0.57\%$\downarrow$)} & 61.29 & 62.13& \\
Training-free (\tiny{Ins}) & 92.62 \tiny{(0.57\%$\downarrow$)} & 37.10 & 38.37 & \\
\midrule
Triggerless & 89.18 \tiny{(4.26\%$\downarrow$)} & 58.47 & 55.22 & \\
\midrule
w/o. Contrastive & 92.53 \tiny{(0.67\%$\downarrow$)} & 82.26 & 81.74 \\
w/o. $L_{claim}$ & 92.95 \tiny{(0.21\%$\downarrow$)} & 80.24 & 79.55 \\
\midrule
\model{} & 92.34 \tiny{(0.87\%$\downarrow$)} & 82.26 & 82.57 \\
\bottomrule
\end{tabular}
\end{table}

%% file: Tables/eval_w_defense_gpt2.tex
\begin{table}[t]
\centering
\caption{Backdoor attack results against GPT2 on the Fake News dataset with defense methods.}
\label{tab:eval_with_defense_gpt2}
\scriptsize
{
\begin{tabular}{@{}l@{\hspace{7pt}}cccc@{}}
\toprule
 & \multicolumn{2}{c}{\textbf{Word-based (\tiny{T1})}} & \multicolumn{2}{c}{\textbf{\model{}}} \\
\cmidrule(lr){2-3} \cmidrule(lr){4-5}
& \textbf{MiASR} & \textbf{MaASR} & \textbf{MiASR} & \textbf{MaASR} \\
\midrule
{RAP}    & 78.40 \tiny{(10.80$\downarrow$)} & 62.38 \tiny{(10.30$\downarrow$)} & 84.32 \tiny{(5.58$\downarrow$)} & 81.29 \tiny{(5.96$\downarrow$)} \\
{STRIP}  & 72.82 \tiny{(16.38$\downarrow$)} & 56.41 \tiny{(16.27$\downarrow$)} & 89.20 \tiny{(0.70$\downarrow$)} & 84.02 \tiny{(3.23$\downarrow$)} \\
{DAN}    & 72.13 \tiny{(17.07$\downarrow$)} & 58.86 \tiny{(13.82$\downarrow$)} & 58.19 \tiny{(31.71$\downarrow$)} & 68.06 \tiny{(19.19$\downarrow$)} \\
\bottomrule
\end{tabular}
}
\end{table}

%% file: Tables/eval_cluster.tex
\begin{table}[t]
\centering
\caption{Backdoor attack results against BERT on the Fake News dataset using different clustering methods.}
\label{tab:eval_cluster}
\scriptsize
\begin{tabular}{lcccc}
\toprule
\textbf{Clustering} & & \textbf{CACC} & \textbf{MiASR} & \textbf{MaASR} \\
\midrule
\multirow{2}{*}{DBSCAN} & \tiny{Benign} & 97.04 & - & - \\
& \tiny{Compromised} & 96.27 \tiny{(0.79\%$\downarrow$)} & 88.50 & 85.05 \\
\midrule
\multirow{2}{*}{MeanShift} & \tiny{Benign} & 97.01 & - & - \\
& \tiny{Compromised} & 96.48 \tiny{(0.55\%$\downarrow$)} & 90.71 & 90.39 \\
\midrule
\multirow{2}{*}{OPTICS} & \tiny{Benign} & 97.09 & - & - \\
& \tiny{Compromised} & 96.31 \tiny{(0.80\%$\downarrow$)} & 88.64 & 85.32 \\
\bottomrule
\end{tabular}
\end{table}

%% file: Tables/eval_roberta.tex
\begin{table*}[t]
\centering
\caption{Backdoor attack results against RoBERTa on three classification datasets.}
\label{tab:eval_roberta}
\scriptsize
\begin{tabular}{@{}l@{\hspace{10pt}}cccc@{\hspace{2pt}}cccc@{\hspace{2pt}}ccc@{}}
\toprule
 & \multicolumn{3}{c}{\textbf{\footnotesize Fake News}} & & \multicolumn{3}{c}{\textbf{\footnotesize Misinformation}} & & \multicolumn{3}{c}{\textbf{\footnotesize Political}} \\ 
\cmidrule{2-4} \cmidrule{6-8} \cmidrule{10-12}
 & \textbf{CACC} & \textbf{MiASR} & \textbf{MaASR} & & \textbf{CACC} & \textbf{MiASR} & \textbf{MaASR} & & \textbf{CACC} & \textbf{MiASR} & \textbf{MaASR} \\ \midrule
Benign & 97.32 & - & - & & 96.16 & - & - & & 87.28 & - & - \\
\midrule
Word-based (\tiny{T1}) & 87.30 \tiny{(10.3\%$\downarrow$)} & 95.47 & 88.96 & & 88.65 \tiny{(7.81\%$\downarrow$)} & 92.30 & 79.35 & & 81.74 \tiny{(6.53\%$\downarrow$)} & 81.74 & 72.21 \\
\midrule
Word-based (\tiny{T2}) & 95.79 \tiny{(1.57\%$\downarrow$)} & 82.58 & 68.97 & & 94.99 \tiny{(1.22\%$\downarrow$)} & 83.50 & 59.31 & & 86.95 \tiny{(0.38\%$\downarrow$)} & 43.40 & 36.60 \\
\midrule
Word-based (\tiny{T3}) & 96.91 \tiny{(0.42\%$\downarrow$)} & 68.64 & 51.24 & & 95.78 \tiny{(0.40\%$\downarrow$)} & 53.06 & 30.27 & & 87.19 \tiny{(0.10\%$\downarrow$)} & 14.47 & 12.77 \\
\midrule
Training-free (\tiny{Sub}) & 90.78 \tiny{(6.72\%$\downarrow$)} & 21.95 & 15.14 & & 92.99 \tiny{(3.30\%$\downarrow$)} & 29.95 & 52.51 & & 84.59 \tiny{(3.08\%$\downarrow$)} & 30.19 & 24.79 \\
\midrule
Training-free (\tiny{Ins}) & 91.02 \tiny{(6.47\%$\downarrow$)} & 20.91 & 18.81 & & 92.05 \tiny{(4.27\%$\downarrow$)} & 17.36 & 18.61 & & 84.78 \tiny{(2.86\%$\downarrow$)} & 30.19 & 25.65 \\
\midrule
Triggerless & 87.00 \tiny{(10.6\%$\downarrow$)} & 34.84 & 23.79 & & 91.01 \tiny{(5.36\%$\downarrow$)} & 21.39 & 48.01 & & 85.37 \tiny{(2.19\%$\downarrow$)} & 23.90 & 25.98 \\
\midrule
w/o. Contrastive & 97.25 \tiny{(0.07\%$\downarrow$)} & 87.11 & 77.18 & & 96.08 \tiny{(0.08\%$\downarrow$)} & 92.30 & 82.91 & & 87.21 \tiny{(0.07\%$\downarrow$)} & 80.50 & 77.82 \\
\midrule
w/o. $L_{claim}$ & 96.88 \tiny{(0.45\%$\downarrow$)} & 87.46 & 80.21 & & 96.05 \tiny{(0.11\%$\downarrow$)} & 89.12 & 93.10 & & 87.29 \tiny{(0.01\%$\uparrow$)} & 83.65 & 82.79 \\
\midrule
Full model & 97.01 \tiny{(0.32\%$\downarrow$)} & 90.24 & 86.03 & & 96.05 \tiny{(0.11\%$\downarrow$)} & 90.34 & 90.18 & & 87.02 \tiny{(0.30\%$\downarrow$)} & 85.53 & 84.23 \\
\bottomrule
\end{tabular}
\end{table*}